\pdfoutput=1

\documentclass[11pt]{article}
\usepackage[dvipsnames]{xcolor}

\usepackage[]{EACL2023}

\usepackage{times}
\usepackage{latexsym}
\usepackage{amsmath}

\usepackage[T1]{fontenc}

\usepackage[utf8]{inputenc}

\usepackage{microtype}

\usepackage{inconsolata}

\usepackage{times}
\usepackage{latexsym}
\usepackage{booktabs}
\usepackage{enumitem}
\usepackage[textsize=scriptsize]{todonotes}

\usepackage{multirow}
\usepackage{subcaption}
\usepackage{dblfloatfix}
\usepackage{amssymb}%
\usepackage{pifont}%
\usepackage{stackengine}

\newcommand{\tablefontsize}{\small \setlength\tabcolsep{4 pt}}

\newcommand\textss[1]{\stackengine{.6ex}{}{\tiny#1}{O}{l}{F}{F}{L}}
\newcommand{\daemark}{\textss{\texttt{D}}}
\newcommand{\emmark}{\textss{\texttt{E}}}

\usepackage{enumitem} 
\newenvironment{enumSpecial}{%
\begin{enumerate}[leftmargin=*, noitemsep]
}
{%
\end{enumerate}
}

\title{Shortcomings of Question Answering Based Factuality Frameworks\\for Error Localization}

\author{Ryo Kamoi \hspace{0.7cm} Tanya Goyal \hspace{0.7cm} Greg Durrett \\
        Department of Computer Science \\
        The University of Texas at Austin \\
        {\tt ryokamoi@utexas.edu}
}

\begin{document}
\maketitle

\begin{abstract}
Despite recent progress in abstractive summarization, models often generate summaries with factual errors. Numerous approaches to detect these errors have been proposed, the most popular of which are question answering (QA)-based factuality metrics. These have been shown to work well at predicting summary-level factuality and have potential to localize errors within summaries, but this latter capability has not been systematically evaluated in past research. In this paper, we conduct the first such analysis and find that, contrary to our expectations, QA-based frameworks fail to correctly identify error spans in generated summaries and are outperformed by trivial exact match baselines. Our analysis reveals a major reason for such poor localization: questions generated by the QG module often inherit errors from non-factual summaries which are then propagated further into downstream modules. Moreover, even human-in-the-loop question generation cannot easily offset these problems. Our experiments conclusively show that there exist fundamental issues with localization using the QA framework which cannot be fixed solely by stronger QA and QG models.

\end{abstract}

\section{Introduction}
Although abstractive summarization systems \citep{Rush2015ASummarization, see-etal-2017-get, Lewis2019BART:Comprehension} have improved drastically over the past few years, these systems often introduce factual errors into generated summaries \citep{Cao2018FaithfulSummarization, kryscinski-etal-2019-neural}. Recent work has proposed a number of approaches to detect these errors, including using off-the-shelf entailment models \citep{falke-etal-2019-ranking,laban-etal-2022-summac}, question answering (QA) models \citep{Chen_Wu_Wang_Ding_2018, wang-etal-2020-asking, durmus-etal-2020-feqa}, and discriminators trained on synthetic data \citep{kryscinski-etal-2020-evaluating}. Such methods have also been explored to identify error spans within summaries \citep{Goyal2020EvaluatingEntailment} and perform post-hoc error correction \citep{dong-etal-2020-multi, cao-etal-2020-factual}.

\begin{figure}[t]
    \centering
    \includegraphics[width=\columnwidth,trim=170px 545px 990px 110px,clip]{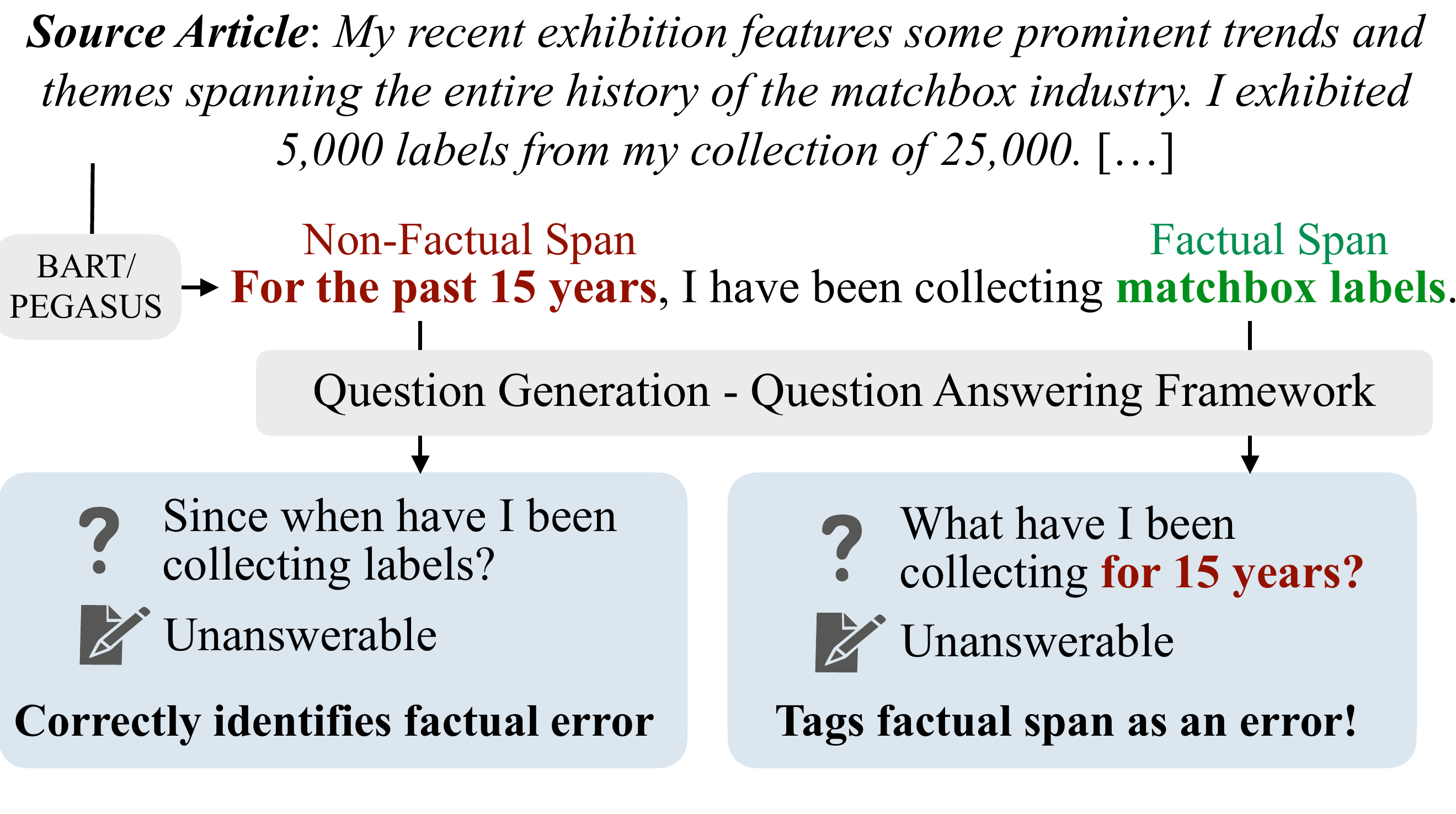} %
    \caption{Factual error localization using QA metrics. Questions are generated for summary spans and then answered by a QA model using the source article as context. For factual spans (e.g. \textit{matchbox labels}), we expect the predicted answers to match the original spans. However, non-factual spans in generated questions inherited from summaries may render these unanswerable and lead to incorrect error localization.}
    \label{fig:first-figure}
\end{figure}

Among these different approaches for evaluating factuality, QA-based frameworks are the most widely adopted \citep{Chen_Wu_Wang_Ding_2018, scialom-etal-2019-answers, durmus-etal-2020-feqa, wang-etal-2020-asking, scialom-etal-2021-questeval, fabbri-etal-2022-qafacteval}. These evaluate the factuality of a set of spans in isolation, then combine them to render a summary-level judgment. Figure \ref{fig:first-figure} illustrates the core mechanism: question generation (QG) is used to generate questions for a collection of summary spans, typically noun phrases or entities, which are then compared with those questions' answers based on the source document to determine factuality. Due to this span-level decomposition of factuality, QA frameworks are widely believed to localize errors \citep{Chen_Wu_Wang_Ding_2018, wang-etal-2020-asking, gunasekara-etal-2021-using-question}. Therefore, the metrics have been applied in settings like post-hoc error correction \citep{dong-etal-2020-multi}, salient \citep{Deutsch2021Question-BasedSummarization} and incorrect \citep{scialom-etal-2021-questeval} span detection, and text alignment \citep{Weiss2021QA-Align:Propositions}. However, their actual span-level error localization performance has not been systematically evaluated in prior work.

\begin{figure*}[t]
    \centering
    \includegraphics[width=\textwidth,trim=100px 400px 350px 320px,clip]{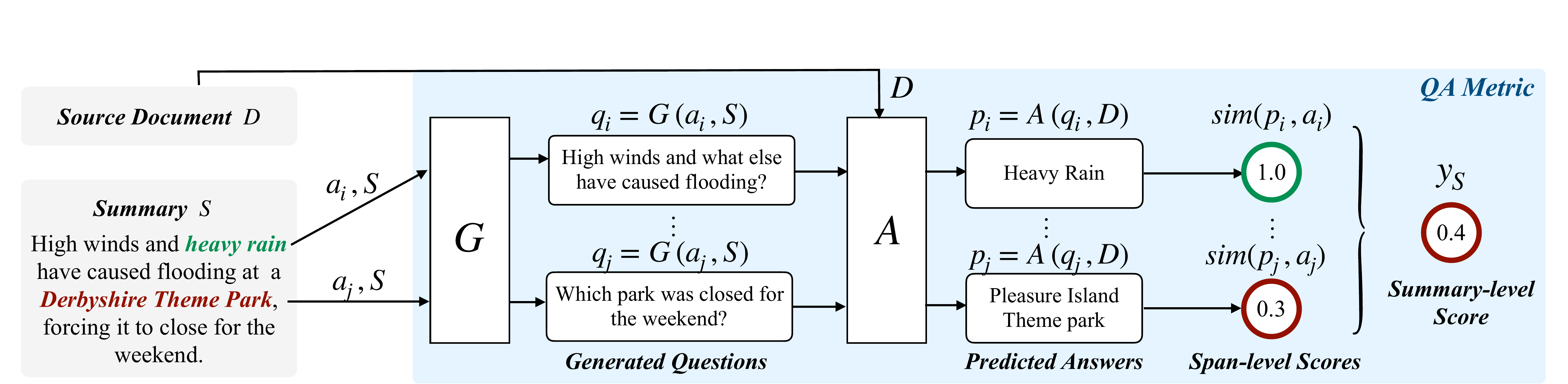}
    \caption{Overall workflow for the QA metrics. First, questions are generated for all NEs and NPs in the generated summary. Answers to these questions are obtained from the source document. Then, a factuality score is computed for each summary span based on it similarity with the predicted span from the previous step. Finally, all span-level scores are aggregated to obtain the final summary-level factuality.}
    \label{fig:qa-based-metric}
\end{figure*}

In this paper, we aim to answer the following question: \textbf{does the actual behavior of QA-based metrics align with their motivation?} Specifically, we evaluate whether these models successfully identify error spans in generated summaries, independent of their final summary-level judgment. We conduct our analysis on two recent factuality datasets \citep{cao-wang-2021-cliff, goyal-durrett-2021-annotating} derived from pre-trained summarization models on two popular benchmark datasets: CNN/DM \citep{hermann2015teaching, nallapati-etal-2016-abstractive} and XSum \citep{narayan-etal-2018-dont}. Our results are surprising: \textbf{we find that good summary-level performance is rarely accompanied by correct span-level error detection.} Moreover, even trivial exact match baselines outperform QA metrics at error localization. Our results clearly show that although motivated by span-level decomposition of the factuality problem, the actual span-level predictions of QA metrics are very poor.

Next, we analyze these failure cases to understand why QA-based metrics diverge from their intended behavior. We find that the most serious problem lies in the question generation (QG) stage: generated questions for non-factual summaries inherit errors from the input summaries (see Figure~\ref{fig:first-figure}). This results in poor localization wherein factual spans get classified as non-factual due to presupposition failures during QA.
Furthermore, we show that such inherited errors cannot be easily avoided: decreasing the length of generated questions reduces the number of inherited errors, but very short questions can be under-specified and not provide enough context for the QA model. %
In fact, replacing automatic QG with human QG also does not improve the error localization of QA metrics. These results demonstrate fundamental issues with the current QA-based factuality frameworks that cannot be patched by stronger QA/QG methods.

Our contributions are as follows. (1) We show that QA-based factuality models for summarization exhibit poor error localization capabilities. (2) We provide a detailed study of factors in QG that hamper these models: inherited errors in long generated questions and trade-offs between these and short under-specified questions. (3) We conduct a human study to illustrate the issues with the QA-based factuality framework independent of particular QA or QG systems. %

\section{QA-Based Factuality Metrics}
Recent work has proposed numerous QA-based metrics for summarization evaluation, particularly factuality \citep{Chen_Wu_Wang_Ding_2018, scialom-etal-2019-answers, eyal-etal-2019-question, durmus-etal-2020-feqa, wang-etal-2020-asking, Deutsch2021Question-BasedSummarization}. These proposed metrics follow the same basic framework (described in Section~\ref{sec:qa-metrics-framework}), and primarily differ in the choice of off-the-shelf models used for the different framework components (discussed in Section~\ref{qa-metrics-compared}).

\subsection{Basic Framework} 
\label{sec:qa-metrics-framework}
Given a source document $D$ and generated summary $S$, the QA-based metrics output a summary-level factuality score $y_{S}$ that denotes the factual consistency of $S$. This includes the following steps (also outlined in Figure~\ref{fig:qa-based-metric}):
\begin{enumSpecial}
    \item \textbf{Answer Selection}: First, candidate answer spans $a_i \in S$ are extracted. These correspond to the base set of \textit{facts} that are compared against the source document $D$. Metrics evaluated in this work \cite{scialom-etal-2021-questeval, fabbri-etal-2022-qafacteval} consider all noun phrases and named entities in generated summaries as the answer candidates set, denoted by $span(S)$.
    \item \textbf{Question Generation}: Next, a question generation model ($G$) is used to generate questions for these answer candidates with the generated summary $S$ as context. Let $q_i = G(a_i, S)$ denote the corresponding question for span $a_i$.
    \item \textbf{Question Filtering}: Questions for which the question answering ($A$) model's predicted answer $A(q_i, S)$ from the summary does not match the original span $a_i$ are discarded, i.e., when $a_i \neq A(q_i, S)$. This step is used to ensure that the effects of erroneous question generation do not percolate down the pipeline; however, answer spans that do not pass this phase cannot be evaluated by the method.
    \item \textbf{Question Answering}: For each generated question $q_i$, the $A$ model is used to predict answers using the source document $D$ as context. Let $p_i = A(q_i, D)$ denote the predicted answer. 
    \item \textbf{Answer Comparison}: Finally, the predicted answer $p_i$ is compared to the expected answer $a_i$ to compute a similarity score $sim(p_i, a_i)$. The overall summary score $y_{S}$ is computed by averaging over all span-level similarity scores: %
    \begin{equation*} \label{eq:qa-summary-level-evaluation}
    y_{S} = \frac{1}{|span(S)|} \sum_{a_i \in span(S)}{sim(A(q_i, D), a_i)} %
    \end{equation*}
\end{enumSpecial}
Based on the motivation behind QA metrics, these similarity scores $sim(p_i, a_i)$ should indicate the factuality of the corresponding spans. If span $a_i$ is factual, then the $G-A$ pipeline should output $p_i \in D$ with high similarity to $a_i$. Conversely, if $a_i$ is non-factual, the similarity score $sim(p_i, a_i)$ should be low. While prior research has only evaluated their sentence-level performance, we use these span-level factuality scores to additionally evaluate the localization performance of QA metrics.

\subsection{QA Metrics compared}
\label{qa-metrics-compared}
In this work, we focus our analysis on the two best performing QA-based metrics from prior work:

\paragraph{QuestEval (QE)} \citet{scialom-etal-2021-questeval} generate questions for answer spans extracted from both the summary (``precision questions'') and source  document (``recall questions''). We only use the former in our experiments as these are shown to correlate better with factuality. Both the $A$ and $G$ components of QuestEval use T5-Large models \citep{Raffel2020ExploringTransformer} fine-tuned on question answering datasets \citep{rajpurkar-etal-2018-know, trischler-etal-2017-newsqa}. The similarity score $sim(p_i, a_i)$ in this framework is computed as the average of the lexical overlap, BERTScore, and the answerability score predicted by $A$.
 
\paragraph{QAFactEval (QAFE)} \citet{fabbri-etal-2022-qafacteval} conduct an ablation study over the different combinations of available $A$ and $G$ models. Here, we use their best-performing combination: an \textsc{Electra}-based $A$ model and a \textsc{Bart}-based $G$ model fine-tuned on the QA2D dataset \citep{Demszky2018TransformingDatasets}. The $sim(p_i, a_i)$ score is obtained using the learned metric LERC \citep{chen-etal-2020-mocha}. If $A(q_i, D)$ is unanswerable for span $a_i$, QAFactEval sets the similarity score $sim(\_, a_i) = 0$ instead of using the LERC metric.

\section{Experimental Setup}
\subsection{Task Definition}
\label{sec:task-definition}
Given document $D$ and a generated summary $S$, let $y_S^* \in \{0, 1\} $ denote the gold summary-level factuality label. Additionally, we assume access to  $L = \{(a, y_{a}^*)\}$ which denotes the set of spans $a \in span(S)$ and their corresponding  span-level gold factuality labels $y_{a}^* \in \{0, 1\}$. 

First, we evaluate the \textbf{summary-level performance} of factuality models, i.e., is the predicted factuality equal to the gold factuality judgment $y_S^*$?  To do this, we covert the predicted factuality score $y_S$ to a binary judgment using dataset-specific thresholds. For each factuality model evaluated, we select thresholds that yield the best F1 scores on the validation set on each dataset.

Next, we evaluate the \textbf{span-level (localization) performance} of factuality models. Similar to the previous setting, we convert span-level predictions $y_{a}$ to binary labels using the best-F1 threshold derived from the validation set.
We report the macro-averaged performance at correctly predicting the span-level label $y_{a}^* \; \forall a\in span(S)$ across all $(D, S)$ pairs in the evaluation dataset. 

To align with the current QA frameworks, we restrict our evaluation to spans that correspond to named entities and noun phrases. This takes a generous view of the QA metrics' performance as it does not penalize them for failing to identify factual-errors outside NPs and NEs. This setting allows us to study the fundamental issues with the QA framework instead of those that can potentially be addressed by extending the question types considered in the framework.

Note that even for NP and NE spans, sometimes the QA metric does not return a span-level prediction if the span has not been selected as an answer candidate or has been discarded during the question filtering phase. We assume the predicted label $y_a = 1$ for such spans, as the model failed to detect any errors.\footnote{Operationally, we set $sim = 6.0$ for QAFactEval and $sim = 1.0$ for QuestEval for filtered spans.} We discuss the performance loss due to this additional filtering step in Appendix~\ref{appendix:un-evaluated-spans}.

\subsection{Datasets}
We conduct our analysis on two human-annotated factuality datasets from prior work that provide gold annotations of factuality at the token level. To the best of our knowledge, these two are the only datasets that include span-level factuality annotation for summaries generated by SOTA models.

\textbf{CLIFF} \citep{cao-wang-2021-cliff} is a dataset consisting of summaries generated by \textsc{Bart} \citep{Lewis2019BART:Comprehension} and \textsc{Pegasus} \citep{Zhang2020PEGASUS:Summarization} models trained on the XSum and CNN/DM summarization datasets. For each generated summary, the dataset includes token-level factuality labels $y_t^* \in \{0, 1\}$. For $y_t^*=0$, these are additionally labeled with fine-grained error types: extrinsic, intrinsic, or requiring world knowledge.

\textbf{GD21} \citep{goyal-durrett-2021-annotating} contains XSum summaries generated using a fine-tuned \textsc{Bart} model. Similar to CLIFF, it contains token-level factuality labels for all generated summaries. 

\paragraph{Deriving gold summary- and span-level factuality labels from human annotations} %
To derive the summary-level gold label $y_S^*$ from these token-level human annotations, we set $y_S^* = 1$ iff all tokens are factual, i.e. $y_t^* = 1 \; \forall t \in S$. %
To derive span-level gold labels, for each NP/NE span $a$, we set $y_a^* = 1$ iff all tokens $t \in a$ are factual.

\begin{table}[t]
    \centering
    \tablefontsize
    \begin{tabular}{c|c|c|cc}
    \toprule
    \textbf{Label} & \multirow{2}{*}{\textbf{Metric}} & \textbf{GD21} & \multicolumn{2}{c}{\textbf{CLIFF}} \\
    \textbf{Gran.} & & XSum & C/D & XSum \\
    \midrule
\multirow{2}{*}{\textbf{Summ.}} & Total & $46$ & $150$ & $150$ \\
 & \% Non-Factual & $52.2$ & $15.3$ & $70.7$ \\ \midrule
\multirow{2}{*}{\textbf{Span}} & \# per summary & $7.9$ & $15.3$ & $5.4$ \\
 & \% Non-Factual & $9.9$ & $1.9$ & $28.1$ \\ \midrule
 \multirow{3}{*}{\textbf{Token}} & \# per summary & $17.1$ & $31.6$ & $13.1$ \\
 & \% Non-Factual & $8.8$ & $1.7$ & $24.5$ \\
 & \% Ignored (Non-Factual) & $28.9$ & $24.5$ & $27.6$ \\
    \bottomrule
    \end{tabular}
    \caption{Test set statistics for CLIFF and GD21 at different levels of label granularity. All our evaluation is done at the summary- and span-levels to align with the QA metrics' formulation. We convert the token-level human annotations to span-level to achieve this. The table reports the \% of non-factual tokens outside NE/NPs that are ignored by the QA metrics' evaluation pipeline.}
    \label{tab:dataset_stats}
\end{table}

We construct validation and test sets by dividing each dataset into equal subsets. The statistics for the test set are included in Table \ref{tab:dataset_stats}. It shows that \textasciitilde26\% of non-factual tokens do not correspond to NEs or NPs and are therefore ignored by the QA metrics' evaluation pipeline. Also, note that the error statistics differ for the XSum summaries in GD21 and CLIFF due to the differences in the annotation methodologies and the trained models used (both \textsc{Bart} and \textsc{Pegasus} in CLIFF vs only \textsc{Bart} in GD21). 

\begin{figure*}[t]
    \centering
    \includegraphics[scale=0.40, trim=0mm 190mm 270mm 0mm,clip]{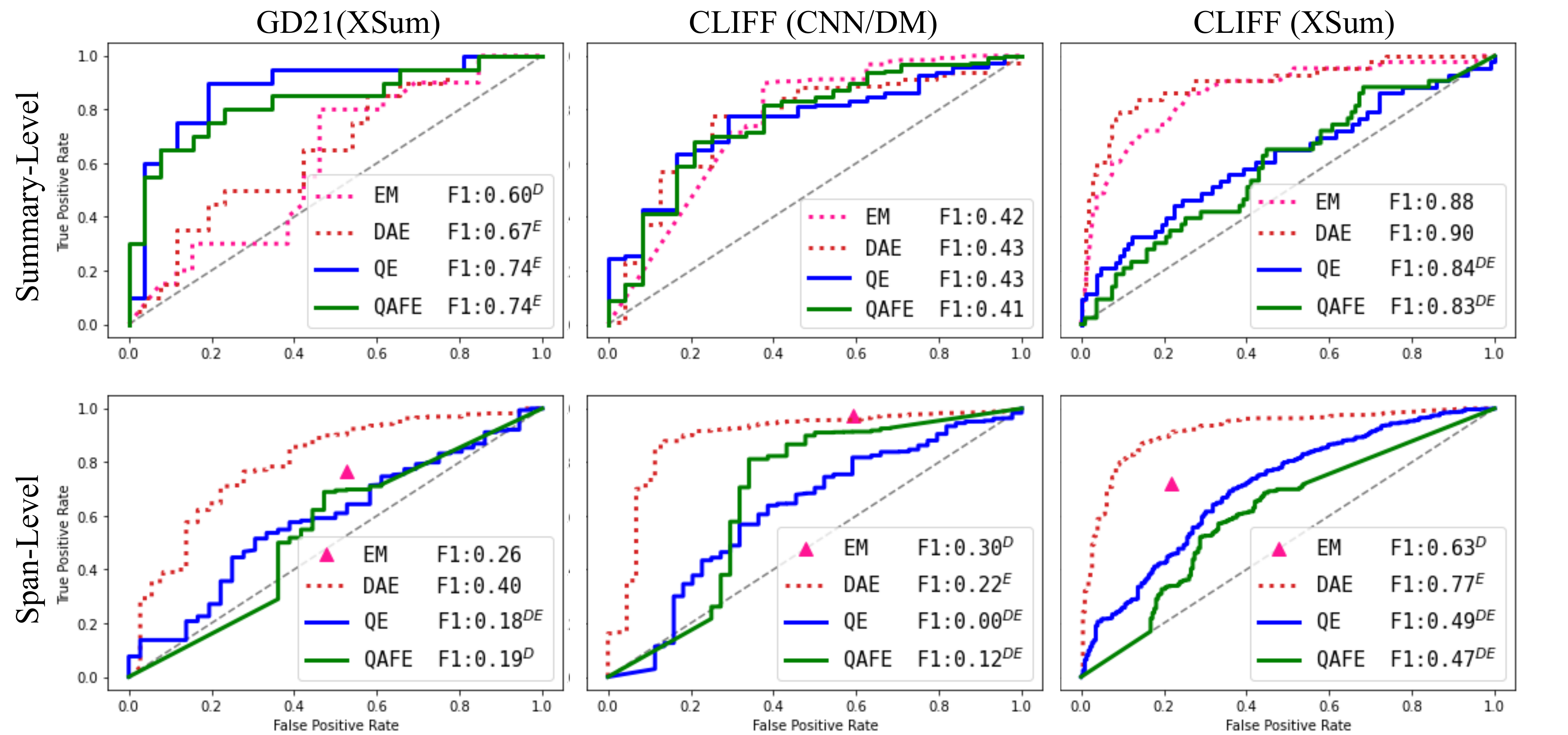}
    \caption{ROC curves and F1 scores (legend) for all systems at the summary- and span-levels (EM is a point as it provides hard binary judgments). Here, $\daemark$ (or $\emmark$) denotes that the performance difference with DAE (or EM) is statistically significant according to a paired bootstrap test (p-value $< 0.05$). We observe that for QA metrics, good summary-level performance (e.g. on GD21 and CLIFF (CNN/DM)) does not imply good localization performance.}
    \label{fig:roc-curves}
\end{figure*}

\subsection{Baselines for Comparison}
\paragraph{Exact Match Baseline (EM)} first extracts all nouns, proper nouns, numbers, adjectives, and pronoun tokens from the generated summary $S$. For these tokens, we set $y_t = 1$ if $y_t \in D$, else $y_t = 0$. We use the fraction of tokens predicted as factual as a summary-level score.

\paragraph{Dependency-Arc Entailment (DAE)}
\citet{Goyal2020EvaluatingEntailment} evaluate the factuality of each dependency arc in generated summaries separately. We follow the methodology proposed by \citet{goyal-durrett-2021-annotating} to derive both summary- and token-level factuality scores from these arc-level judgments. We refer readers to the original paper for further details. We use their available model checkpoint in our experiments.\footnote{Code and trained model checkpoint provided by authors at: \url{https://github.com/tagoyal/factuality-datasets}}

We convert token-level judgments from these baseline models into span-level judgments to make their outputs compatible with our evaluation framework. This is described in detail in Appendix~\ref{sec:convert-token-to-span}.

\section{Summary vs.~Span Level Performance}%
\label{sec:span-vs-summary}
QA metrics motivate the use of span-level factuality as building blocks for evaluating factuality at the sentence level. Therefore, \textbf{our hypothesis is that good summary-level performance must be accompanied by good span-level performance}. Here, we test this by comparing summary- and span-level performances.

Figure~\ref{fig:roc-curves} outlines the performance of QA metrics and baseline systems. The top row shows ROC curves and F1 scores (in the legend) for all three datasets; the bottom row shows span-level results. The dotted black lines show the performance of a random baseline. First, we observe that none of the baselines or QA metrics have a clear advantage over other systems for all datasets at the summary-level. For instance, QA metrics outperform baselines on GD21, and show similar performance on CLIFF (CNN/DM) and worse performance on the CLIFF (XSum) dataset. However, \textbf{across all datasets, we see that there exists a substantial mismatch between the performance of QA metrics at the summary- and span-levels.} Notably, for GD21, both QE and QAFE substantially outperform baseline models at the summary-level, but exhibit much poorer span-level performance. Similarly, QA metrics are comparable to baselines at the summary-level for CLIFF (CNN/DM) but much worse at the span-level. On the other hand, the error localization performance of the DAE model is more consistent with its summary-level performance. Surprisingly, the trivial \textbf{exact match (EM) baseline consistently outperforms QA metrics at error localization for all datasets}. These results clearly show that our hypothesis is false: QA-based metrics do not provide reliable span-level explanations for their summary-level predictions. 

Note that the diagonal lines in the span-level ROC curves for QA metrics arise due to a large number of spans being assigned the same factuality scores. As discussed in Section~\ref{sec:qa-metrics-framework}, some spans are filtered during the question filtering stage (Step 3) if their corresponding generated questions are of low quality. We consider these to be factual and assign them the maximum factuality score; this results in the diagonal line from $(0, 0)$.\footnote{Note that QE generates multiple questions for each span and therefore rarely discards spans (it is not likely that all questions are bad). Therefore, it has a shorter diagonal line.} We study the effects of this span filtering on localization performance in Appendix~\ref{appendix:un-evaluated-spans}. Additionally, QAFE assigns the same factuality score ($=0$) to all spans with unanswerable questions resulting in the diagonal line to $(1, 1)$. 

\section{Why do QA metrics fail at span-level error localization?}%
\label{sec:extrinsic-error-in-questions}
Consider the error localization task in the example in Figure~\ref{fig:first-figure}. Here, the QA metric needs to correctly distinguish between the factual span ``\emph{matchbox labels}'' and the extrinsic error ``\emph{for the past 15 years}''. For such summaries (containing a mix of factual and non-factual spans), we observed that the generated questions for factual spans often \textbf{inherit} non-factual summary spans. Given such questions, e.g. ``\emph{What have I been collecting for 15 years?}'', an ideal QA model \emph{\textbf{should}} predict unanswerable (even though that hurts localization) as the source article does not include any mention of an item being collected for 15 years. Based on this observation, we hypothesize that such inherited errors in generated questions adversely affect the localization performance of automatic QA metrics by misclassifying factual spans.

\begin{figure}[t]
    \centering
    \includegraphics[width=\columnwidth,trim=150px 420px 940px 240px,clip]{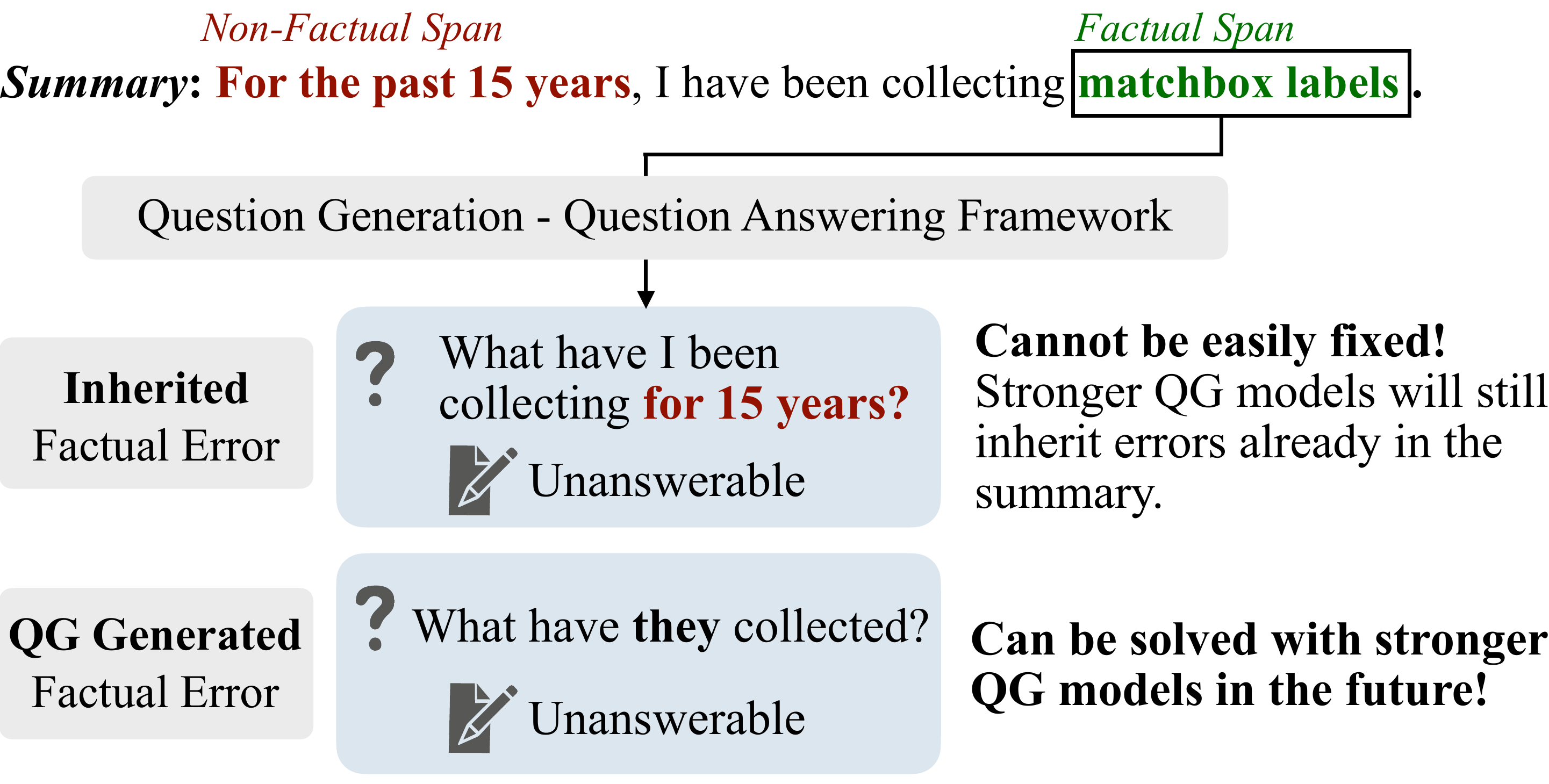}
    \caption{Generated questions broadly include two types of errors: (1) inherited errors that copy non-factual spans from the summary, and (2) errors introduced by imperfect QG models. While this latter set of errors may be eliminated by stronger QG models, inherited errors cannot be easily fixed.
    }
    \label{fig:inherited_errors}
\end{figure}

First, we draw a clear distinction between (1) errors inherited from summaries, and (2) those introduced due to generation errors by the QG model. Figure \ref{fig:inherited_errors} illustrates these two separate cases. We note that the latter set of errors can potentially be addressed by stronger QG models in the future. Our analysis in this section only studies the former set, i.e. inherited errors, as these will persist independent of the improvement in QA/QG models.

\paragraph{What percentage of questions are impacted by inherited errors?} First, we determine the scope of the limitation introduced by inherited errors. Table~\ref{tab:error_in_question_ratio} outlines how frequently generated questions contain inherited errors; we report these numbers only for the non-factual summaries as only these are affected by inherited errors. We define errors in a summary as inherited when a question copies at least one token that is annotated as non-factual. We use error type labels (extrinsic or intrinsic) present in the CLIFF and GD21 datasets to report separate numbers for these phenomena\footnote{Generated questions can inherit both types of errors. In the tables in this section, ``extrinsic error'' denotes questions that inherit at least one extrinsic error, but ``only intrinsic error'' denotes questions that only inherit intrinsic errors.}. For the CLIFF dataset, we include world knowledge errors within the extrinsic type. In general, we observe that inherited errors are more common in QAFactEval compared to QuestEval; this can be attributed to the longer length questions generated by the former (see Appendix~\ref{appendix:qa-based-metrics-question-stats} for details).

\begin{table}[t]
    \centering
    \tablefontsize
    \begin{tabular}{c|c|ccc}
    \toprule
    \multirow{2}{*}{\textbf{QA Metric}} & \textbf{Type of} & \textbf{GD21} & \multicolumn{2}{c}{\textbf{CLIFF}} \\
    & \textbf{Inherited Error} & XSum & C/D & XSum \\
    \midrule
    \multirow{2}{*}{QuestEval} &
extrinsic error & $19.2$ & $9.1$ & $44.8$ \\
 & 
only intrinsic error & $25.7$ & $17.3$ & $3.6$ \\
    \midrule
    \multirow{2}{*}{QAFactEval} &
extrinsic error & $39.1$ & $11.1$ & $93.1$ \\
& 
only intrinsic error & $48.9$ & $34.2$ & $6.0$ \\
    \bottomrule
    \end{tabular}
    \caption{
        Percentage of questions that inherit extrinsic and intrinsic errors from summaries. We only consider non-factual summaries, i.e., summaries containing at least one non-factual span in this table. 
    }
    \label{tab:error_in_question_ratio}
\vskip 1em
    \centering
    \tablefontsize
    \begin{tabular}{c|c|ccc}
    \toprule
    \multirow{2}{*}{\textbf{QA Metric}} & \textbf{Type of} & \textbf{GD21} & \multicolumn{2}{c}{\textbf{CLIFF}} \\
    & \textbf{Inherited Error} & XSum & C/D & XSum \\
    \midrule
    \multirow{3}{*}{QuestEval} &
extrinsic error & $\phantom{0}3.1$ & $ 93.9$ & $ 30.5$ \\
 & 
only intrinsic error & $\phantom{0}9.3$ & $ 97.3$ & $ 50.0$ \\
 & 
no inherited error & $15.5$ & $ 98.7$ & $ 56.0$ \\
    \midrule
    \multirow{3}{*}{QAFactEval} &
extrinsic error & $\phantom{0}7.7$ & $ 65.4$ & $ 63.8$ \\
& 
only intrinsic error & $29.2$ & $ 82.0$ & $ 40.0$ \\
& 
no inherited error & $29.3$ & $ 92.8$ & $ 65.4$ \\
    \bottomrule
    \end{tabular}
    \caption{
        Percentage of factual spans correctly classified by QA metrics, i.e. $y_t = y^*_t = 1$.  We use the same thresholds as for F1 scores in Figure~\ref{fig:roc-curves}. Results show that inherited errors lead to more erroneous classification as non-factual across all datasets.
    }
    \label{tab:error_caused_by_error_in_question}
\end{table}

\paragraph{Do inherited errors in generated questions hurt factuality prediction?} To answer this, we zoom in on factual spans in generated summaries (we consider \emph{both} factual and non-factual summaries here), and investigate how often these are erroneously classified as non-factual. We report this for three different scenarios: (1) w/ inherited extrinsic error, (2) w/ inherited intrinsic errors only, and (3) w/o any inherited error.  
Table~\ref{tab:error_caused_by_error_in_question} outlines our results. We observe that across all settings, factual spans with inherited errors in their corresponding questions are more likely to be erroneously classified as non-factual compared to those with no inherited errors. Between error types, we observe that extrinsic inherited errors tend to harm localization more than intrinsic errors. 

Note that inherited errors are only observed for summaries that are \emph{already non-factual}. Therefore, erroneous classification of factual spans as non-factual hurts span-level but does \emph{not} hurt summary-level performance. In fact, \citet{fabbri-etal-2022-qafacteval} show that longer questions (which typically inherit more extrinsic errors, but do not cause summary-level error) exhibit better summary-level performance compared to shorter questions (which can be under-specified and cause summary-level error). This indicates that there exists a trade-off in performance between these different granularity levels. 

\begin{figure}[t]
    \centering
    \includegraphics[width=\columnwidth,trim=85mm 82mm 290mm 80mm,clip]{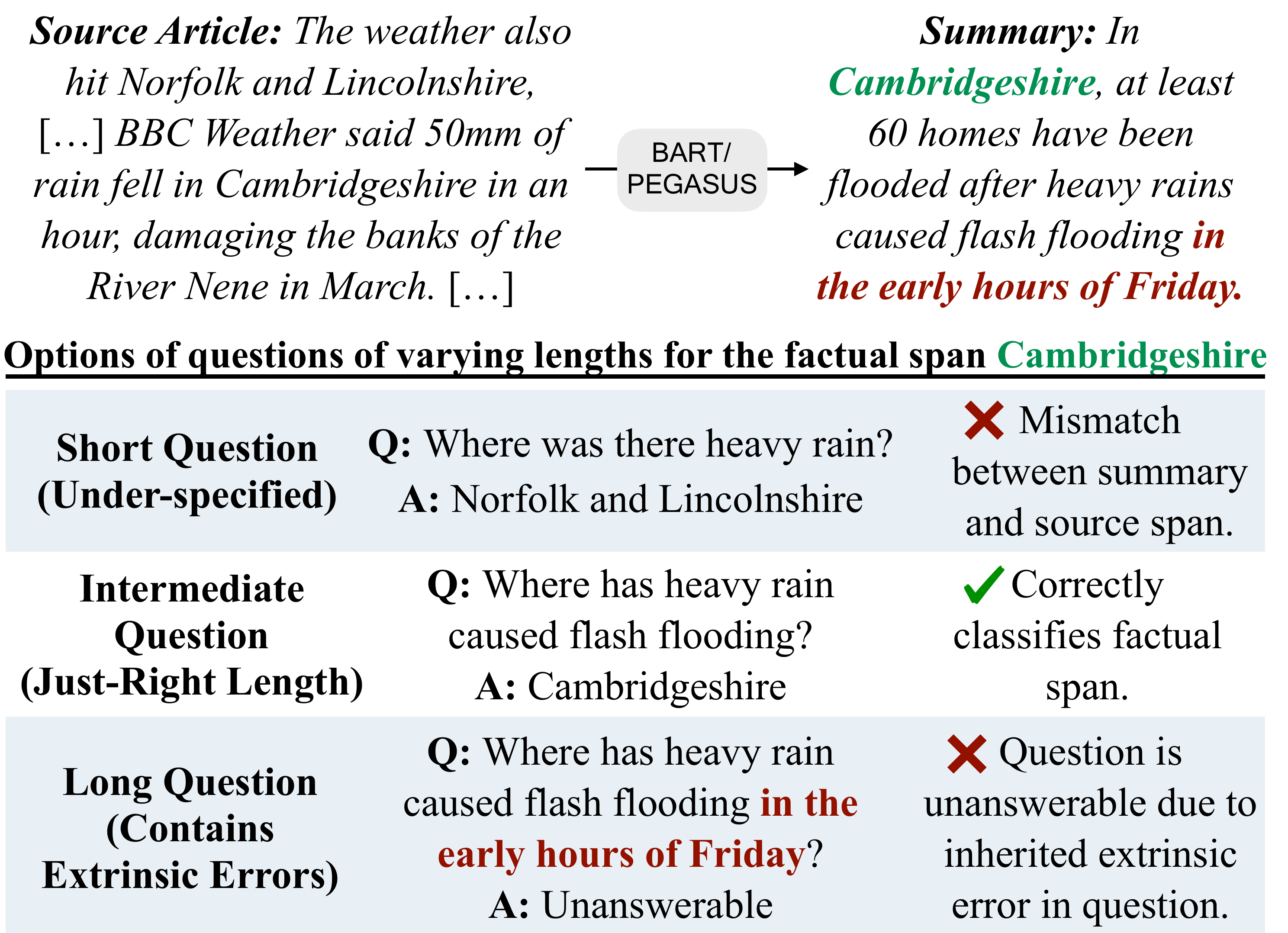}
    \caption{Specificity and lengths of generated questions affect the answers from QA models and confuse span-level evaluation. While ``\emph{Cambridgeshire}'' is a factual span, questions with inappropriate specificity can cause QA models to make mistakes. It is impossible to know what length of question is ``just right'' during question generation.}
    \label{fig:question-lengths}
\end{figure}

\paragraph{Can we avoid inherited errors in generated summaries?} Since we do not have prior knowledge of which spans in summaries contain factual errors, we cannot trivially ensure that generated questions do not inherit the errors. One possible strategy could be to generate very short questions that include minimal details from the summary to avoid inheriting non-factual spans from the summary. However, these may then suffer from being too under-specified. We illustrate this in Figure~\ref{fig:question-lengths}. Consider the factual span ``\emph{Cambridgeshire}''. The shortest question in the figure ``\emph{Where was there heavy rain?}'' is not under-specified for the summary, since it is the only place name in the summary. However, there are multiple possible answers in the source document, and QA models may \emph{reasonably} answer ``\emph{Norfolk and Lincolnshire}'', leading to erroneous classification of ``\emph{Cambridgeshire}'' as non-factual. Therefore, {\bf there exists a trade-off between under-specified (short) questions and over-specified (long) questions} and it is difficult to predict the optimal level of specificity. This problem cannot be addressed by improving QA models; an ideal QA model will return unanswerable to questions with inherited errors and will be still confused by under-specified questions. We explore this issue further using human question generation in Section~\ref{sec:human-QG}.

\section{Can Human QG Improve Localization?}%
\label{sec:human-QG}
In Section~\ref{sec:extrinsic-error-in-questions}, we discussed how the number of inherited errors can be indirectly influenced by varying the length of generated questions and the accompanying trade-offs: longer questions are more likely to inherit errors but shorter questions may be under-specified. Here, we investigate this using perfect QG, i.e., replacing automatic QG with humans. We evaluate two aspects: (1) How does question length impact localization? (2) Does human QG improve localization?

\subsection{Experiment Design}

For each summary and candidate span pair $(S, a_i)$, we obtain human-written questions of varying lengths and information content.\footnote{Question lengths could also be varied if we used distinct automatic QG models, but by choosing human QG, we avoid conflating the impact of varying question specificity/length with errors or other performance differences in models.} Then, we replace the QG module of QAFactEval with these human-written questions to study the effect of question length on error localization performance.\footnote{We also considered using human QA; however, we found that the QA task is ill-defined for humans when questions themselves contain extrinsic errors. \citet{fabbri-etal-2022-qafacteval} also show that QA performance has less impact on factuality.}

Annotators generate 3 types of questions: %
\begin{enumSpecial}
    \item \textbf{Shortest possible question} such that given the question-summary pair, humans can unambiguously identify the correct span in the summary.
    \item \textbf{Longest question} incorporating as much information from the generated summary as reasonably allowed, often including almost an entire summary sentence.
    \item \textbf{Intermediate questions} with levels of information content between the above two extremes. We allow annotators to generate any number of such intermediate questions. 
\end{enumSpecial}

\paragraph{Annotation and Setup} We conduct this experiment on 150 randomly selected summaries from the CLIFF dataset. For the CNN/DM subset, we only selected non-factual summaries, since CNN/DM contains a small number of non-factual spans. Human annotators manually generated 2,186 questions (please refer to Appendix~\ref{appendix:human-qg} for details).\footnote{Human generated questions are provided at: \url{https://github.com/ryokamoi/QA-metrics-human-annotation}} We use half of the summaries as validation sets.

We evaluate localization performance using 4 different length configurations for human-written questions: short, long, intermediate, and oracle. For intermediate questions, evaluation is always done over three questions. We randomly sub-sample (or over-sample) from this set if more (or fewer) than three are available and report their average performance. If no intermediate question is written, we randomly sample from the other two categories. This only happens when the length difference between the shortest and longest questions is small. For the oracle setting, we report results using the question for each span that leads to the best localization performance. In other words, we use the highest scoring question for factual spans and the lowest for non-factual spans.

\subsection{Results}

\begin{figure}[t]
    \centering
    \includegraphics[width=\columnwidth]{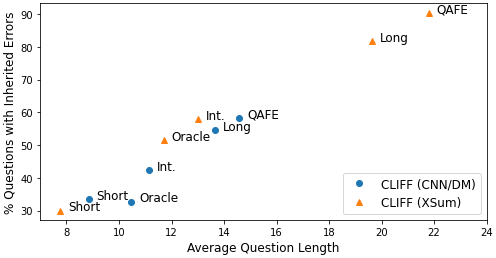}
    \caption{
        Statistics for questions generated by human annotators and QAFactEval (QAFE). ``\% Questions with Inherited Errors'' is the percentage of questions that inherit non-factual spans from non-factual summaries. As expected, longer questions are more likely to inherit factual errors from the generated summaries.
    }
    \label{fig:human-question-length}
\end{figure}

Figure~\ref{fig:human-question-length} outlines statistics for questions generated by human annotators and the QG model of QAFactEval (QAFE) generated for the same spans. As expected in Section~\ref{sec:extrinsic-error-in-questions}, it shows that the percentage of questions that inherit non-factual spans in summaries increases with length. In this figure, we only analyze non-factual summaries since questions generated for factual summaries do not inherit errors. This result verifies our assumption and shows that we can analyze a trade-off between long questions that tend to inherit more non-factual spans from summaries and short questions with fewer inherited errors but can be under-specified.

\begin{figure}[!t]
    \centering
    \includegraphics[width=\columnwidth,trim=0 750px 1180px 0, clip]{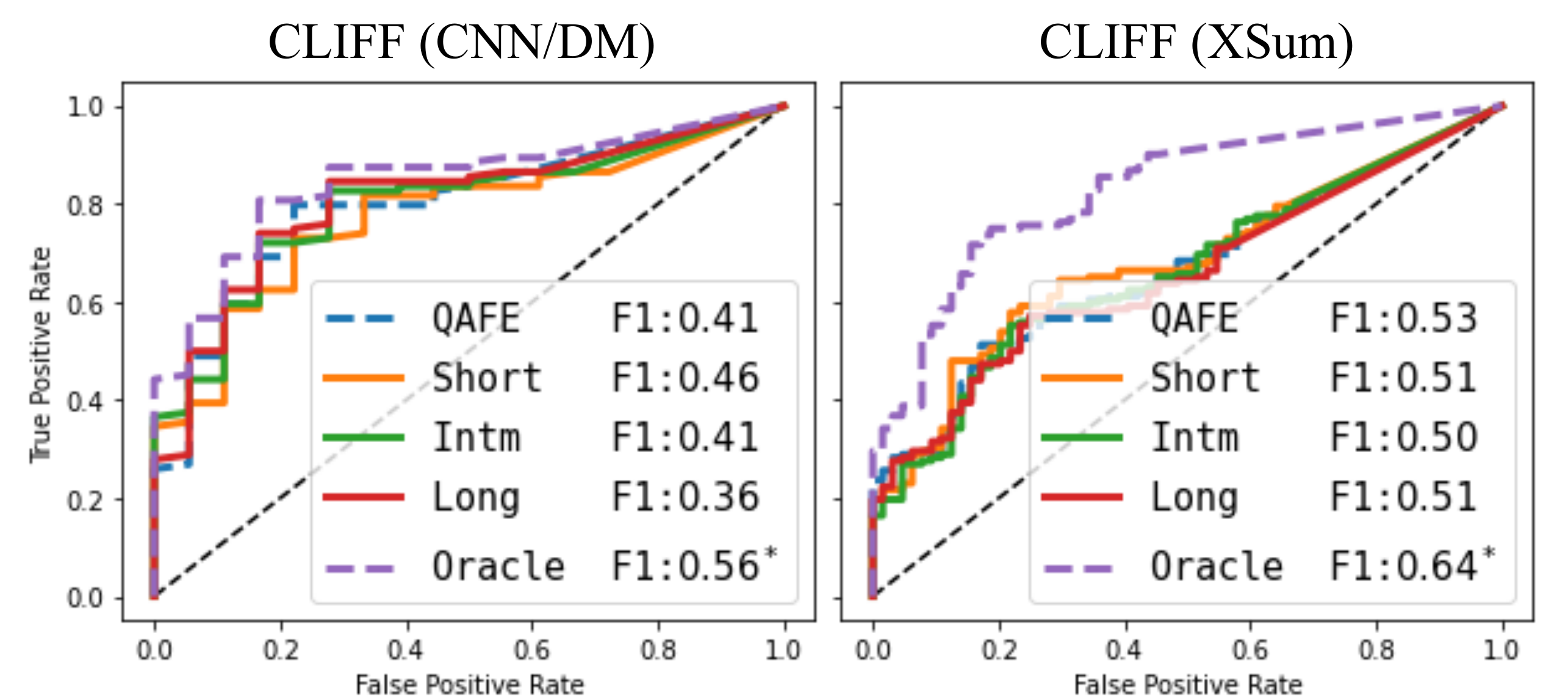}
    \caption{ROC curves and F1 scores (in legend) for span-level performances using human-written questions. These results show that no single question length configuration (except oracle) can outperform automatic QG. $^*$ denotes that improvement over QAFactEval is statistically significant (paired bootstrap test, p-value $< 0.05$).}
    \label{fig:human_generated_questions_rocs}
\end{figure}

\paragraph{Error Localization} Figure~\ref{fig:human_generated_questions_rocs} outlines the span-level localization performance for these different human question configurations and the QG model of QAFactEval. First, we notice that \textbf{human QG does not improve the localization performance of the QA frameworks}, with all three configurations exhibiting similar performance to the fully automatic QAFactEval (QAFE) model. \textbf{However, the \textit{oracle} questions report significant improvement over QAFactEval}; this indicates that while there does exist an optimal length question for most spans, there isn't a clear pattern that can help select it during evaluation. We again note that it is not possible to select an optimal question length for each span without prior knowledge about errors in summaries. We conclude that the overall failure of human QG to improve over QAFactEval suggests that there exist fundamental issues with the QA-based factuality formulations which cannot be simply fixed by stronger QG models.

\section{Discussion}
Analysis in our paper suggests that QA-based metrics have fundamental problems which will be difficult to address in future work. \textbf{Our view is that future system designers should favor entailment-based approaches \citep{falke-etal-2019-ranking,laban-etal-2022-summac} as a result.} One reason for this is that successful QA-based approaches actually implement something similar to entailment. Both our analysis and \citet{fabbri-etal-2022-qafacteval} show that we can improve summary-level performance by generating long questions to avoid underspecified questions. However, answering questions that contain almost all the information about a sentence can be regarded as a weak form of entailment evaluation: it assesses whether the question-answer pairs that include all information about the sentence are entailed by the original document. Compared to entailment, the answer comparison step can introduce difficulties and long questions may still lead to incorrect evaluation. Since this paper shows that QA-based metrics do not lead to interpretable, localizable judgments about errors, QA-based metrics do not seem to have any structural advantage over entailment-based metrics.

\section{Related Work} %
Recent work \citep{fabbri-etal-2021-summeval} has shown that popular metrics like
ROUGE \citep{Lin2004ROUGE:Summaries} and BERTScore \citep{Zhang2019BERTScore:BERT}
correlate poorly with human judgments of summary quality. Factuality, in particular, has been widely studied in recent summarization literature \cite{kryscinski-etal-2019-neural, falke-etal-2019-ranking}, both from the perspective of identifying non-factual generations \citep{wang-etal-2020-asking, durmus-etal-2020-feqa, Goyal2020EvaluatingEntailment} and improving the factuality of summarization models themselves \citep{kang-hashimoto-2020-improved, cao-wang-2021-cliff}.

The majority of the work in factual evaluation has focused on summary-level metrics and is not capable of localizing errors within summaries. Recent work has decomposed factuality into summaries' dependency arcs \citep{Goyal2020EvaluatingEntailment} or semantic-graph representations \citep{ribeiro2022factgraph}. These localization capabilities have several downstream applications like post-editing \citep{zhao-etal-2020-reducing, Chen2021ImprovingSelection}, removing noisy training data \citep{nan-etal-2021-entity, goyal-durrett-2021-annotating}, among others. 

\section{Conclusion} %
In this work, we show that although QA-based factuality metrics are motivated by error localization, in practice, they exhibit extremely poor localization capabilities. We provide a detailed analysis of the different issues in current metrics that hinder better localization performance. Finally, we run a human study to investigate whether human-level QG can fix some of these issues and conclude that there exist fundamental issues with the QA framework that cannot be simply fixed by stronger models. 

\section{Limitations}
Given the lack of prior study in error localization of summarization evaluation, there is no large-scale dataset with token-level or span-level factuality labels. Constantly-evolving summarization models also mean that any such dataset would be come outdated in a fairly short time. However, we believe that the fundamental issues we discussed with QA metrics would persist across different summarization model outputs, despite our evaluation over a limited set.

Note that all our analysis is conducted on English language datasets and models of summarization, with a limited focus on newswire summaries. We believe that the issues identified here will transfer to other languages, but other domains such as dialogue or narrative summaries may exhibit substantially different types of factuality errors. These have not been studied as heavily in prior work, so likely new techniques and analysis will be needed for these settings.

\section*{Acknowledgments}

We thank Juan Diego Rodriguez for helpful comments on this work. This work was partially supported by NSF grant IIS-2145280, a gift from Salesforce Research, and a gift from Amazon. The authors acknowledge the Texas Advanced Computing Center (TACC) at the University of Texas at Austin for providing HPC resources used to conduct this research.

\bibliography{references,custom,anthology}
\bibliographystyle{acl_natbib}

\appendix
\section{Additional Implementation Details}

\subsection{QuestEval}
We use version 0.2.4 of the implementation by the authors with the recommended parameters (\texttt{task = 'summarization'}, \texttt{ do\_weighter = False}).\footnote{https://github.com/ThomasScialom/QuestEval}

\subsection{DAE}
Our experiments use a trained model provided by authors (\texttt{DAE\_xsum\_human\_best\_ckpt}).
This model is trained on a subset of XSum dataset \citep{Maynez2020OnSummarization} that is distinct from examples in CLIFF and GD21.

\subsection{Converting Tokens-level factuality scores to span-level scores} \label{sec:convert-token-to-span}

For the EM and the DAE baselines provide token-level factuality scores. We refer readers to the original DAE paper for details on how token-level scores are obtained \citep{goyal-durrett-2021-annotating}. However, all our evaluation is designed to be at the span-level to align with QA metrics. To convert token-level scores to span-level, we annotate a span as non-factual if it contains any non-factual token. 

\subsection{Statistics for Ignored non-NP/NE tokens}
The QA metrics do not evaluate the factuality of any token outside the boundary of a named entity of a noun phrase (discussed in Section~\ref{sec:task-definition}). In Table~\ref{tab:ignored_pos}, we show which kinds of tokens are ignored by the QA metrics but annotated as non-factual in our human annotated factuality datasets. Figure~\ref{tab:ignored_tokens_example} provides an illustrative example of such ignored non-factual tokens in the different datasets).

\begin{table}[h]
    \centering
    \small
    \begin{tabular}{crrr}
    \toprule
    \multirow{2}{*}{} & GD21 & \multicolumn{2}{c}{CLIFF} \\
     & XSum & C/D & XSum \\
    \midrule
adposition & $30.8$ & $25.3$ & $29.9$ \\
verb & $24.4$ & $23.7$ & $23.5$ \\
auxiliary & $15.2$ & $10.0$ & $17.1$ \\
punctuation & $12.6$ & $21.2$ & $17.5$ \\
particle & $6.7$ & $3.9$ & $4.5$ \\
    \bottomrule
    \end{tabular}
    \caption{Statistics for non-factual POS-tags ouside the NP/NE boundaries and ignored by the QA metrics.}
    \label{tab:ignored_pos}
\end{table}

\begin{table}[!t]
    \centering
    \tablefontsize
    \begin{tabular}{p{0.95\columnwidth}}
        \toprule
        \multicolumn{1}{c}{\bf CLIFF (XSum)} \\
An environmental permit {\bf \color{red} has} {\bf \color{red} been} {\bf \color{red} revoked} {\bf \color{red} following} a fire {\bf \color{red} at} a fuel recycling plant {\bf \color{red} in} Manchester{\bf\color{red}.} 
        \\
        \midrule
        \multicolumn{1}{c}{\bf CLIFF (CNN/DM)} \\
A Japan Railway maglev train {\bf \color{red} hit} 603 kilometers {\bf \color{red} per} hour{\bf \color{red} (}374 miles {\bf \color{red} per} hour{\bf \color{red})} {\bf \color{red} on} an experimental track {\bf \color{red} in} Yamanashi Tuesday{\bf\color{red}.} A spokesperson {\bf \color{red} said} the train {\bf \color{red} spent} 10.8 seconds {\bf \color{red} traveling} {\bf \color{red} above} 600 km {\bf \color{red} per} hour {\bf \color{red} ,} {\bf \color{red} during} which it {\bf \color{red} covered} 1.8 kilometers {\bf \color{red} (} 1.1 miles {\bf \color{red} )} That {\bf \color{red} 's} nearly 20 football fields {\bf \color{red} in} the time it {\bf \color{red} took} you {\bf \color{red} to} {\bf \color{red} read} the last two sentences {\bf \color{red} .} Japan Railways {\bf \color{red} has} {\bf \color{red} been} {\bf \color{red} testing} their train {\bf \color{red} to} {\bf \color{red} figure} {\bf \color{red} out} the best operational speed {\bf \color{red} for} a planned route {\bf \color{red} between} Tokyo {\bf \color{red} and} Nagoya {\bf \color{red} scheduled} {\bf \color{red} to} {\bf \color{red} begin} service {\bf \color{red} in} 2027 {\bf \color{red} .} 
        \\
        \midrule
        \multicolumn{1}{c}{\bf GD21} \\
high winds {\bf \color{red} and} heavy rain {\bf \color{red} have} {\bf \color{red} caused} flooding {\bf \color{red} at} a derbyshire theme park{\bf\color{red},} {\bf \color{red} forcing} it {\bf \color{red} to} {\bf \color{red} close} {\bf \color{red} for} the weekend{\bf\color{red}.} 
        \\
        \bottomrule
    \end{tabular}
    \caption{
        Example of non-factual tokens outside NP/NE boundaries and ignored by the QA metric in factuality evaluation.
    }
    \label{tab:ignored_tokens_example}
\end{table}

\begin{table}[t]
    \centering
    \tablefontsize
    \setlength{\tabcolsep}{3pt}
    \begin{tabular}{cccc|ccc}
        \toprule
        \multirow{2}{*}{} 
        & \multicolumn{3}{c|}{Avg Question Length}
        & \multicolumn{3}{c}{Avg No.~Questions} \\
        \cmidrule(rl){2-4} \cmidrule(rl){5-7}
        \multirow{2}{*}{} & GD21 & \multicolumn{2}{c|}{CLIFF} & 
        GD21 & \multicolumn{2}{c}{CLIFF} \\
         & XSum & C/D & XSum & XSum & C/D & XSum \\
        \midrule
        QAFactEval & $26.3$ & $16.2$ & $21.1$ & $4.9$ & $10.0$ & $3.9$ \\
        QuestEval & $16.4$ & $11.6$ & $13.9$ & $8.3$ & $20.5$ & $7.5$ \\
        \bottomrule
    \end{tabular}
    \caption{Statistics for the generated questions for the QuestEval and QAFactEval metrics.}
    \label{tab:qa-based-metrics-stats}
\end{table}

\section{Statistics of Generated Questions in QA-Based Metrics} \label{appendix:qa-based-metrics-question-stats}
Table~\ref{tab:qa-based-metrics-stats} provides statistics for the generated questions from QuestEval and QAFactEval, highlighting the difference between these two metrics. On average, QAFactEval generates much longer questions. On the other hand, QuestEval generates a larger number of questions as it often generates multiple questions per candidate span. 

\section{Performance Loss due to Span Filtering} \label{appendix:un-evaluated-spans}
In Section~\ref{sec:task-definition}, we discussed that the current QA metrics do not evaluate non-NP/NE spans. These operational shortcomings prevent these metrics from providing a complete picture of error localization over all summary tokens. Here, we discuss another similar issue arising due to the question filtering step of the overall workflow (Step 3).

\begin{table}[t]
    \centering
    \small
    \begin{tabular}{c|l|ll}
    \toprule
        \multirow{2}{*}{\textbf{Model}} & \textbf{GD21} & \multicolumn{2}{c}{\textbf{CLIFF}} \\
        & XSum & CNN/DM & XSum \\ \midrule
EM & $0.30$ & ${\bf 0.27 }$ & $0.64\daemark$  \\
DAE & ${\bf 0.32 }$ & $0.20\emmark$ & ${\bf 0.78 }$  \\ \midrule
QE & $0.19\daemark\emmark$ & $0.06\daemark\emmark$ & $0.45\daemark\emmark$  \\
QAFE & $0.21\daemark$ & $0.13\daemark\emmark$ & $0.49\daemark\emmark$  \\
    \bottomrule
    \end{tabular} %
    \caption{Span-level performance (F1 scores) over the subset of NP/NEs that are not discarded by either of the two QA metrics. $\daemark$ (or $\emmark$) denotes that the performance difference with DAE (or EM) is statistically significant according to a paired bootstrap test (p-value $< 0.05$). Even under this generous setting, we observe that the QA metric show very poor performance.} %
    \label{tab:filtered-f1}
\end{table}

Although QA metrics select all NP/NE spans for evaluation during the candidate selection stage (Step1), some of these are filtered out if their corresponding question is of low quality: $a_i$ for which $A(q_i, S) \neq a_i$ are also discarded from further evaluation. Since no errors are detected in these spans, they are considered to be factual. 

We observed that this question filtering step removes around 30\% of the NE/NPs in the QAFactEval framework. This implies that this metric only evaluates 70\% of the valid spans, potentially missing factual errors in the remaining NP/NEs. Note that these numbers are considerably lower for QuestEval ($<$5\%) as it generates multiple questions for each candidate span and hence is more likely to include an acceptable question. 

As this impacted the results in Figure~\ref{fig:roc-curves}, we can ask what is the performance of the QA metrics over spans that they \textbf{actually} evaluate for factuality? If this performance is high, we can reasonably assume that the QA metrics' localization capabilities  can be improved through better question generation models. Table~\ref{tab:filtered-f1} outlines our results: we report F1 scores at the span-level when evaluating over the subset of NP/NEs that are evaluated by both the QE and QAFE models. Although the QA metrics report improved results over those reported in Figure~\ref{fig:roc-curves}, these are still low enough so as to not be useful for error localization in practical settings. 

\begin{figure}[!t]
    \centering
    \begin{subfigure}[t]{0.49\columnwidth}
        \centering
        \includegraphics[width=\textwidth]{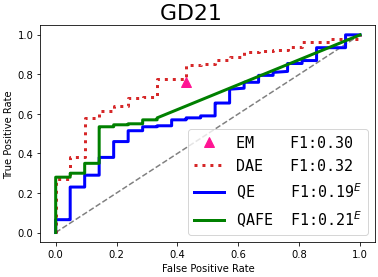}
    \end{subfigure}
    \begin{subfigure}[t]{0.49\columnwidth}
        \centering
        \includegraphics[width=\textwidth]{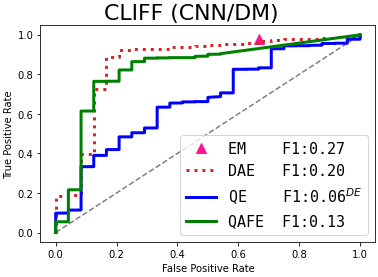}
    \end{subfigure}
    \begin{subfigure}[t]{0.49\columnwidth}
        \centering
        \includegraphics[width=\textwidth]{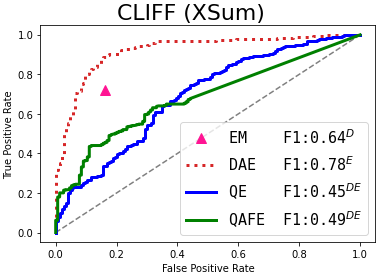}
    \end{subfigure}
    \caption{ROC curves for span-level performance on the subset of NE/NPs evaluated by all the QA metrics.}
    \label{fig:roc-curves_removenone}
\vskip 1em
    \centering
    \begin{subfigure}[t]{0.47\columnwidth}
        \centering
        \includegraphics[width=\textwidth]{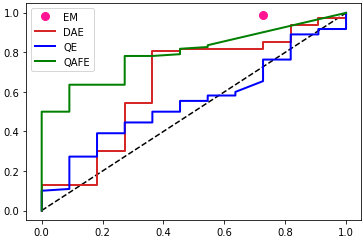}
        \caption{Bottom 50\% of the summaries according to error rate}
    \end{subfigure}
    \hfill
    \begin{subfigure}[t]{0.47\columnwidth}
        \centering
        \includegraphics[width=\textwidth]{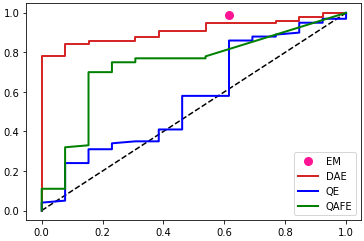}
        \caption{Top 50\% of the summaries according to error rate}
    \end{subfigure}
    \caption{Comparison of ROC curves for the span-level performance on CLIFF (CNN/DM). There shows results on the subset of NP/NEs actually judged by QA metrics. The graphs show that the performance of the QAFE is better on the subset of summaries with a smaller number of errors.}
    \label{fig:roc-cnndm-after-removing-nonevaluated-analysis}
\end{figure}

Figure~\ref{fig:roc-curves_removenone} shows the corresponding ROC curves for these. These show similar trends: the performance of QAFactEval improves when evaluating on this subset, but is still not better than the baseline models. 
Interestingly, for CLIFF (CNN/DM), we found that most of this improvement comes from the subset of candidate spans whose questions contain only a small fraction of factual errors (Figure~\ref{fig:roc-cnndm-after-removing-nonevaluated-analysis}). This aligns with our analysis in Section~\ref{sec:extrinsic-error-in-questions} that showed that errors in generated questions inherited from non-factual summaries was one of the major reasons for performance degradation, since questions generated from summaries with small number of errors are expected to inherit fewer errors.

\begin{table}[t]
    \centering
    \small
    \begin{tabular}{c|p{.66\columnwidth}}
    \toprule
        \multirow{3}{*}{{\bf Summary}} & 
        Plans to build a new hospital {\bf \color{red} in Somerset} have been given a {\bf \color{red} £3m} boost by the government. \\
        \midrule
        {\bf Selected Span} & 
        a new hospital \\
        \midrule
        {\bf Shortest} &
        What does the plan propose to build? \\
        \multirow{2}{*}{{\bf Intermediate}} & 
        What does the plan that has been given a boost propose to build? \\
        \multirow{3}{*}{{\bf Longest}} & 
        What does the plan that has been given a {\bf \color{red} £3m} boost by the government propose to build? \\
        \midrule
        \multirow{3}{*}{{\bf QAFactEval}} & 
        What plans to build {\bf \color{red} in Somerset} have been given a {\bf \color{red} £3m} boost by the government? \\
    \bottomrule
    \end{tabular}
    \caption{Example of human-generated questions.}
    \label{tab:human-generated-questions}
\end{table}

\section{Additional Details about Human QG Annotation} \label{appendix:human-qg}

The human annotation in Section~\ref{sec:human-QG} was done by the authors of this paper. They were provided with summaries and extracted answer candidate spans. For spans that were judged to be invalid (e.g. ``\textit{it}''), they were asked to manually discard these spans. For all others, questions of varying lengths and specificity were written. See an example in Table~\ref{tab:human-generated-questions}. To aid in this question writing step, we also provide the corresponding  QAFactEval questions. For the longest questions, we found that annotators often chose to build on these questions albeit with corrections to the structure and grammar of the automatic questions.

Table~\ref{tab:number_of_human_generated_questions} shows the number of summaries, spans, and generated summaries annotated in our human QG experiments. We use half of the summaries as validation sets.

\begin{table}[t]
    \centering
    \begin{tabular}{c|cc}
\toprule
                & CLIFF    & CLIFF  \\
                & (CNN/DM) & (XSum) \\
\midrule
\# Summary       & 30  & 120 \\
\# Span          & 323 & 470 \\
\# Questions     & 737 & 1449 \\
\bottomrule
    \end{tabular}
    \caption{Number of summaries, spans, and generated questions annotated by human QG.}
    \label{tab:number_of_human_generated_questions}
\end{table}

\newcommand{\spansummarytableheader}{
    & \multicolumn{3}{c|}{Summary-Level} 
    & \multicolumn{3}{c}{Span-Level}  \\
    \cmidrule(lr){2-4} \cmidrule(lr){5-7}
    \multirow{2}{*}{Models} 
     & GD21 & \multicolumn{2}{c|}{CLIFF} 
     & GD21 & \multicolumn{2}{c}{CLIFF} \\
     & XSum & C/D & XSum & XSum & C/D & XSum \\
}
\newcommand{\spansummarytableheadernocolumn}{
    \multicolumn{3}{c|}{Summary-Level} 
    & \multicolumn{3}{c}{Span-Level}  \\
    \cmidrule(lr){1-3} \cmidrule(lr){4-6}
    GD21 & \multicolumn{2}{c|}{CLIFF} 
     & GD21 & \multicolumn{2}{c}{CLIFF} \\
    XSum & C/D & XSum & XSum & C/D & XSum \\
}

\newcommand{\datasetcolumn}{GD21 & \multicolumn{2}{c|}{CLIFF}}
\newcommand{\datasetcolumnend}{GD21 & \multicolumn{2}{c}{CLIFF}}
\newcommand{\splitcolumn}{XSum & C/D & XSum}

\begin{figure*}[!t]
    \centering
    \begin{subfigure}[t]{.19\textwidth}
         \centering
         \includegraphics[width=\textwidth]{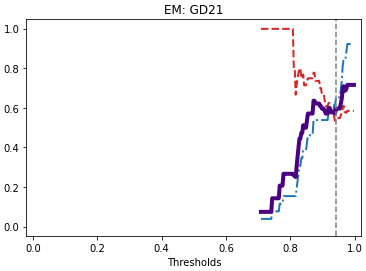}
         \caption{EM Baseline}
    \end{subfigure}
    \begin{subfigure}[t]{.19\textwidth}
         \centering
         \includegraphics[width=\textwidth]{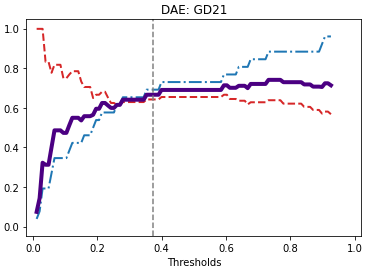}
         \caption{DAE}
    \end{subfigure}
    \begin{subfigure}[t]{.19\textwidth}
         \centering
         \includegraphics[width=\textwidth]{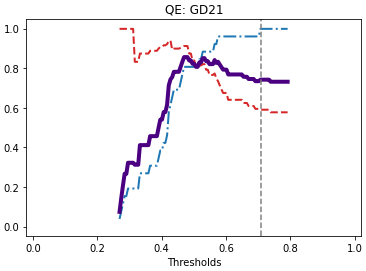}
         \caption{QuestEval}
    \end{subfigure}
    \begin{subfigure}[t]{.19\textwidth}
         \centering
         \includegraphics[width=\textwidth]{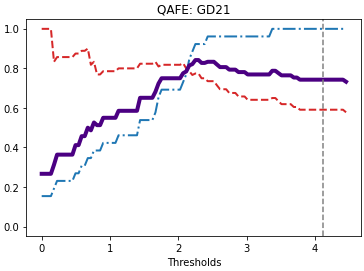}
         \caption{QAFactEval}
    \end{subfigure}
        \caption{Summary-level F1 performance on the GD21 test set at different thresholds}
        \label{fig:summary_f1_thresholds_graph_GD21}
\end{figure*}

\begin{figure*}[!t]
    \centering
    \begin{subfigure}[t]{.19\textwidth}
         \centering
         \includegraphics[width=\textwidth]{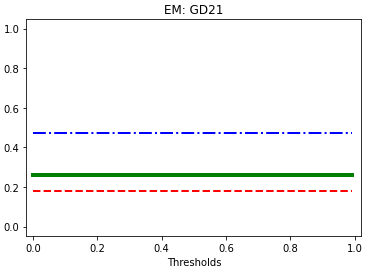}
         \caption{EM Baseline}
    \end{subfigure}
    \begin{subfigure}[t]{.19\textwidth}
         \centering
         \includegraphics[width=\textwidth]{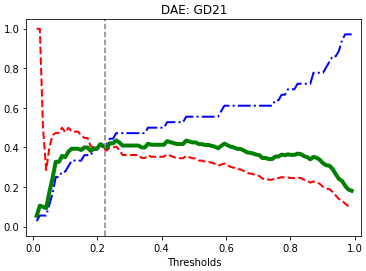}
         \caption{DAE}
    \end{subfigure}
    \begin{subfigure}[t]{.19\textwidth}
         \centering
         \includegraphics[width=\textwidth]{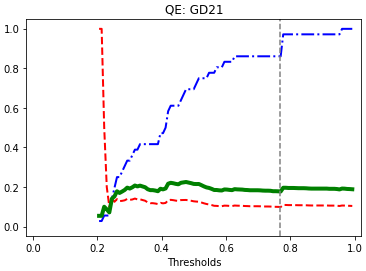}
         \caption{QuestEval}
    \end{subfigure}
    \begin{subfigure}[t]{.19\textwidth}
         \centering
         \includegraphics[width=\textwidth]{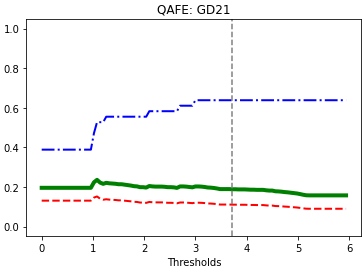}
         \caption{QAFactEval}
    \end{subfigure}
        \caption{Span-level F1 performance on the GD21 test set at different thresholds}
        \label{fig:span_f1_thresholds_graph_GD21}
\end{figure*}

\begin{figure*}[!t]
    \centering
    \begin{subfigure}[t]{.19\textwidth}
         \centering
         \includegraphics[width=\textwidth]{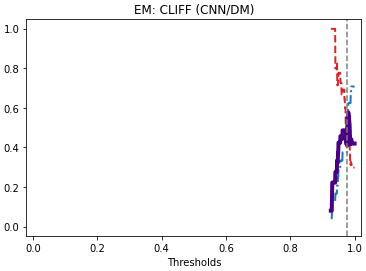}
         \caption{EM Baseline}
    \end{subfigure}
    \begin{subfigure}[t]{.19\textwidth}
         \centering
         \includegraphics[width=\textwidth]{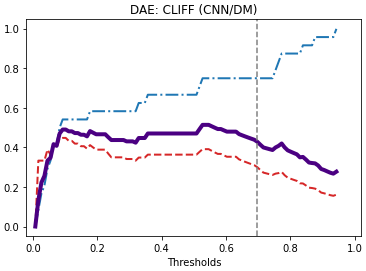}
         \caption{DAE}
    \end{subfigure}
    \begin{subfigure}[t]{.19\textwidth}
         \centering
         \includegraphics[width=\textwidth]{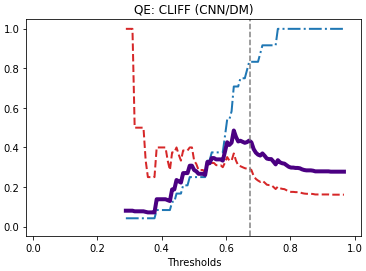}
         \caption{QuestEval}
    \end{subfigure}
    \begin{subfigure}[t]{.19\textwidth}
         \centering
         \includegraphics[width=\textwidth]{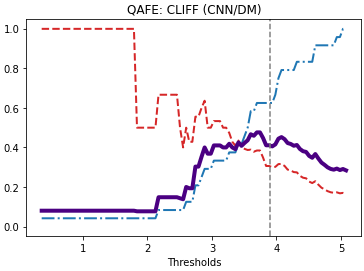}
         \caption{QAFactEval}
    \end{subfigure}
        \caption{Summary-level F1 performance on the CLIFF (CNNDM) test set at different thresholds.}
        \label{fig:summary_f1_thresholds_graph_CLIFF (CNNDM)}
\end{figure*}

\begin{figure*}[!t]
    \centering
    \begin{subfigure}[t]{.19\textwidth}
         \centering
         \includegraphics[width=\textwidth]{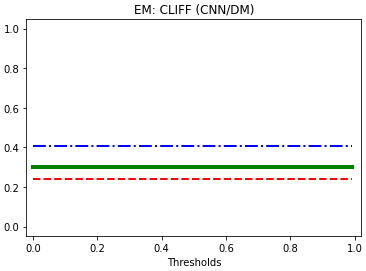}
         \caption{EM Baseline}
    \end{subfigure}
    \begin{subfigure}[t]{.19\textwidth}
         \centering
         \includegraphics[width=\textwidth]{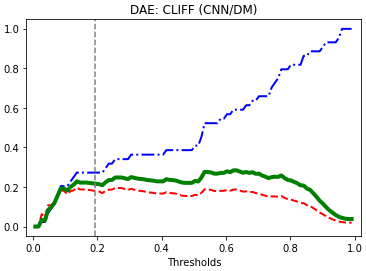}
         \caption{DAE}
    \end{subfigure}
    \begin{subfigure}[t]{.19\textwidth}
         \centering
         \includegraphics[width=\textwidth]{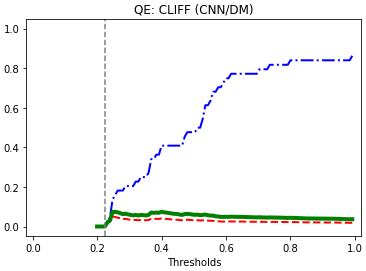}
         \caption{QuestEval}
    \end{subfigure}
    \begin{subfigure}[t]{.19\textwidth}
         \centering
         \includegraphics[width=\textwidth]{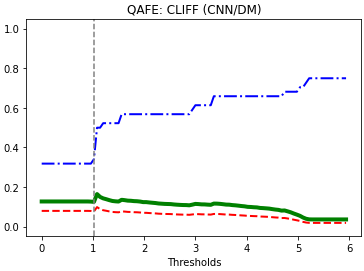}
         \caption{QAFactEval}
    \end{subfigure}
        \caption{Span-level F1 performance on the CLIFF (CNNDM) test set at different thresholds.}
        \label{fig:span_f1_thresholds_graph_CLIFF (CNNDM)}
\end{figure*}

\begin{figure*}[!t]
    \centering
    \begin{subfigure}[t]{.19\textwidth}
         \centering
         \includegraphics[width=\textwidth]{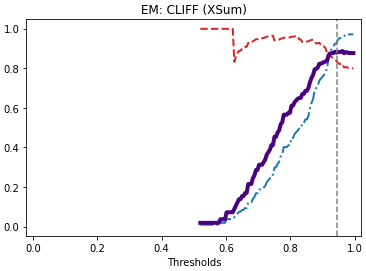}
         \caption{EM Baseline}
    \end{subfigure}
    \begin{subfigure}[t]{.19\textwidth}
         \centering
         \includegraphics[width=\textwidth]{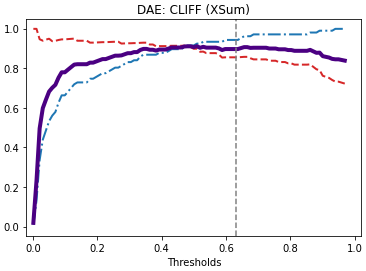}
         \caption{DAE}
    \end{subfigure}
    \begin{subfigure}[t]{.19\textwidth}
         \centering
         \includegraphics[width=\textwidth]{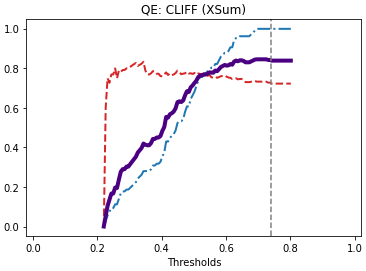}
         \caption{QuestEval}
    \end{subfigure}
    \begin{subfigure}[t]{.19\textwidth}
         \centering
         \includegraphics[width=\textwidth]{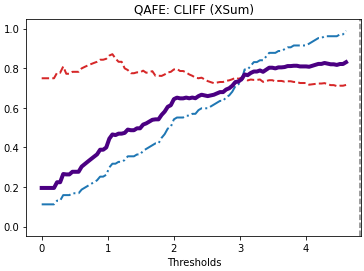}
         \caption{QAFactEval}
    \end{subfigure}
        \caption{Summary-level F1 performance on CLIFF (XSum) test set at different thresholds.}
        \label{fig:summary_f1_thresholds_graph_CLIFF (XSum)}
\end{figure*}

\begin{figure*}[!t]
    \centering
    \begin{subfigure}[t]{.19\textwidth}
         \centering
         \includegraphics[width=\textwidth]{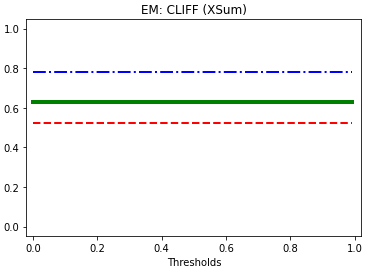}
         \caption{EM Baseline}
    \end{subfigure}
    \begin{subfigure}[t]{.19\textwidth}
         \centering
         \includegraphics[width=\textwidth]{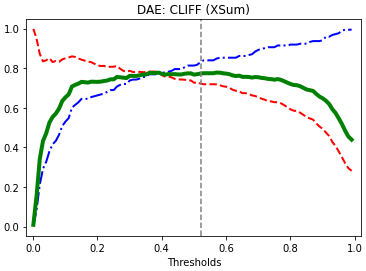}
         \caption{DAE}
    \end{subfigure}
    \begin{subfigure}[t]{.19\textwidth}
         \centering
         \includegraphics[width=\textwidth]{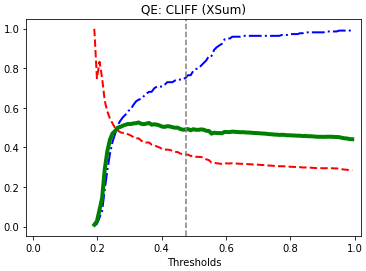}
         \caption{QuestEval}
    \end{subfigure}
    \begin{subfigure}[t]{.19\textwidth}
         \centering
         \includegraphics[width=\textwidth]{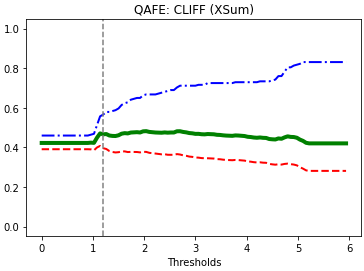}
         \caption{QAFactEval}
    \end{subfigure}
        \caption{Span-level F1 performance on the CLIFF (XSum) test set at different thresholds.}
        \label{fig:span_f1_thresholds_graph_CLIFF (XSum)}
\end{figure*}

\section{Summary and Span Level Evaluation}
\label{appendix:summary-and-span-level-evaluation}
We provide additional results for our experiments in  Figure~\ref{fig:roc-curves}.
Figure~\ref{fig:summary_f1_thresholds_graph_GD21} to \ref{fig:span_f1_thresholds_graph_CLIFF (XSum)} show F1 scores, precision, and recall on test sets with different thresholds.
These show that QA-based metrics cannot yield high precision at any threshold.

\end{document}